\documentclass[
]{ceurart}

\usepackage{listings}
\usepackage{subcaption}
\begin{document}

\copyrightyear{2025}
\copyrightclause{Copyright for this paper by its authors.
  Use permitted under Creative Commons License Attribution 4.0
  International (CC BY 4.0).}


\conference{OVERLAY 2025, 7th International Workshop on Artificial Intelligence and Formal Verification, Logic, Automata, and Synthesis, October 26th, 2025, Bologna, Italy}

\title{RNN Generalization to Omega-Regular Languages}


\author{Charles Pert}[%
orcid=0009-0009-9790-6362,
email=charles.pert@imperial.ac.uk
]
\cormark[1]

\author{Dalal Alrajeh}[%
orcid=0000-0002-1365-8026,
email=dalal.alrajeh@imperial.ac.uk
]

\author{Alessandra Russo}[%
orcid=0000-0002-3318-8711,
email=a.russo@imperial.ac.uk
]

\address{Imperial College London, UK}

\cortext[1]{Corresponding author.}

\begin{abstract}
B\"{u}chi automata (BAs) recognize $\omega$-regular languages defined by formal specifications like linear temporal logic (LTL) and are commonly used in the verification of reactive systems. However, BAs face scalability challenges when handling and manipulating complex system behaviors. As neural networks are increasingly used to address these scalability challenges in areas like model checking, investigating their ability to generalize beyond training data becomes necessary. This work presents the first study investigating whether recurrent neural networks (RNNs) can generalize to $\omega$-regular languages derived from LTL formulas. We train RNNs on ultimately periodic $\omega$-word sequences to replicate target BA behavior and evaluate how well they generalize to out-of-distribution sequences. Through experiments on LTL formulas corresponding to deterministic automata of varying structural complexity, from 3 to over 100 states, we show that RNNs achieve high accuracy on their target $\omega$-regular languages when evaluated on sequences up to $8 \times$ longer than training examples, with $92.6\%$ of tasks achieving perfect or near-perfect generalization. These results establish the feasibility of neural approaches for learning complex $\omega$-regular languages, suggesting their potential as components in neurosymbolic verification methods.
\end{abstract}


\maketitle

\section{Introduction}

\begin{figure}[!ht]
    \centering
    \includegraphics{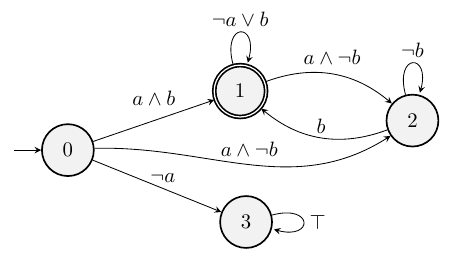}
    \caption{A DBA associated with the LTL formula $\mathbf{G}(a \rightarrow \mathbf{F}b) \land a$ (i.e., whenever $a$ is true, it must eventually be followed by $b$, and $a$ is initially true). Accepting states are double-circled. Transitions are labeled with propositional formulas. Input sequences, $\omega$-words, are composed of symbols that represent assignments to the propositions. For example, the word $(a \land b)^\omega$ is accepted.
    }
    \label{fig:dba}
\end{figure}

Linear Temporal Logic (LTL)~\cite{pnueliLTL} formulas specify properties of system execution traces through $\omega$-regular languages, which are composed of infinitely long sequences called $\omega$-words. While regular languages are recognized by finite automata, $\omega$-regular languages require B\"{u}chi automata (BAs)~\cite{buchi_automata}. See Figure~\ref{fig:dba} for an example of a deterministic B\"{u}chi automaton (DBA). For a detailed introduction to these concepts, we refer the reader to \cite{Baier2008}. 

While BAs provide exact solutions, they can become computationally expensive to handle and manipulate when representing complex behaviors. Recently, neurosymbolic methods have been used in verification contexts such as model checking~\cite{giacobbe2024neural}, areas that traditionally rely on BAs. Characterizing whether neural networks can recognize $\omega$-regular languages is a step toward enabling the development of additional approaches. Recent studies have shown that recurrent neural networks (RNNs) have the ability to generalize to the recognition of regular languages~\cite{butoi2025training,deletang2023neural,svete2024efficiently,merrill-etal-2020-formal}, but this has not yet been shown specifically for $\omega$-regular languages. Related works~\cite{stammet2022analyzing,stammet2023universality} have used graph neural networks to analyze BA properties like emptiness checking and \cite{hahn2021teaching} demonstrated that Transformers~\cite{vaswani2017attention} can generate satisfying $\omega$-words for LTL formulas, yet the problem of generalization to the recognition of $\omega$-regular languages remains open.

Extending the existing work on regular language generalization to $\omega$-regular languages requires handling two practical challenges: (1) encoding: representing infinite sequences with finite-length sequences; (2) labeling sequences: computing acceptance labels for large batches of sequences. While finite automata accept a word when it terminates in an accepting state, BAs require $\omega$-words to traverse accepting states infinitely often, meaning RNNs must learn to recognize eventually periodic behavior instead of just reaching an accepting state. We address challenge (1) by using ultimately periodic (UP) $\omega$-words, which uniquely characterize their $\omega$-regular languages~\cite{ultimately_periodic_result}. This representation enables us to investigate whether RNNs can approximate the symbolic acceptance mechanisms of BAs.

While existing model checking tools like Spot~\cite{spot_tool_citation} can compute acceptance labels for specific sequences, this approach quickly becomes impractical for the large datasets required for neural network training and does not address sequence generation. Instead, we use Spot to construct a BA from each LTL formula, generate sequences using the BA representation and simulate acceptance by directly processing the generated sequences through the BA. We restrict our approach to DBAs to simplify this acceptance check, limiting our study to recurrence properties~\cite{mp_hierarchy}. While this means that this study does not cover persistence properties, recurrence properties cover a large number of properties and are commonly used in verification. Using BAs as data generators provides some control over sampling diversity, which is essential since random sampling can induce data imbalance problems when accepted (or rejected) traces are rare.

This work establishes the feasibility of RNNs generalizing to the recognition of $\omega$-regular languages while identifying challenges for future research. Our investigation complements ongoing neurosymbolic advances in verification, for example, neural certificates for model checking~\cite{giacobbe2024neural}, neural circuit synthesis~\cite{schmitt2021neural}, as well as neural specification mining, learning finite automata or LTL$_f$~\cite{giacomo2013linear} from traces~\cite{luo2025nada,wan2024endtoend,luo2022bridging,walke2021learning}.

The main contributions of this work are:
\begin{itemize}
    \item We present the first empirical evidence that RNNs achieve high accuracy on $\omega$-regular languages when evaluated at lengths up to $8\times$ longer than the sequences in their training distribution.\footnote{Code available at: \url{https://github.com/pertcj/omega-generalization}}
    \item We provide an analysis of RNN generalization, showing that generalization is not limited to toy $\omega$-regular languages and is robust across DBAs with over 100 states. We also show that the model complexity is correlated with the complexity of the BA recognizing the $\omega$-regular language.
\end{itemize}

\section{Method}
\label{sec:method}

Practical training of RNNs on $\omega$-regular languages requires a finite representation of $\omega$-words and an efficient method for computing acceptance labels for generated sequences given the infinite acceptance condition of BAs. Our approach resolves the finite encoding problem by only using UP~$\omega$-words of an $\omega$-regular language, $uv^\omega$, where $u$ is a finite prefix and $v$ is an infinitely repeating suffix; these $\omega$-words uniquely characterize the language~\cite{ultimately_periodic_result}. We encode $uv^\omega$ as $u\$v$~\cite{ultimately_periodic_result}. The alphabet size of a DBA is $2^{|P|} + 1$, where $|P|$ is the number of propositions present in the LTL formula used to construct the DBA. Each symbol in the alphabet represents either an assignment to all propositions or the separator symbol $\$$.

Importantly, the encoding $u\$v$ establishes a bijection between UP~$\omega$-words in the target $\omega$-regular language and words in a derived regular language. The DBA can be algorithmically reconstructed from the finite automaton recognizing this regular language~\cite{ultimately_periodic_result}. Therefore, while our RNNs are learning regular languages (and leveraging their established ability~\cite{deletang2023neural,svete2024efficiently,butoi2025training}), they are learning canonical regular representations of the target DBA that preserve all structural information for reconstruction.

To determine the label of a $u\$v$ sequence, we simulate DBA behavior. Simulating the finite prefix $u$ yields the state in which $u$ terminates. For the suffix $v$, we compute the state transition matrix induced by reading $v$ and use matrix exponentiation to determine reachability. The sequence is accepted if repeated application of $v$ can reach a cycle containing an accepting state. 

We illustrate how we sample $u\$v$ sequences with fixed length $n$ from a DBA. We first sample the position $k$ of $\$$ uniformly between $1$ and $n-1$ and then sample $u$ with length $k-1$ and $v$ with length $n - k$. We sample $u$ and $v$ by uniformly sampling valid paths in the DBA. %
Many DBAs exhibit strong acceptance biases that skew random sampling toward rejection or acceptance. For example, the presence of accepting or rejecting sink states (states that can only transition back to themselves) can dominate uniform sampling. We address this through targeted sampling strategies to improve the balance of our sampled sequences: (1) we oversample sequences, determine their labels, and selectively filter them, targeting a balanced class distribution; (2) when sampling accepted sequences, we exclude transitions to rejecting sink states during sampling (and vice versa for rejected sequences); (3) when sampling rejected sequences, we prevent transitions to accepting states within the suffix $v$ because if $v$ contains a state transition to an accepting state, the resulting $\omega$-word is likely to be accepted. It is still possible to sample equivalent rejected sequences as this constraint is not applied to $u$. By controlling the sampling of accepted and rejected sequences separately, the resulting dataset is more balanced compared to uniform sampling of sequences.

\section{Experiments}
Our experiments aim to answer the following research questions:\begin{description}
    \item\textbf{RQ1}: Can RNNs generalize to the recognition of sequences from $\omega$-regular languages when trained only on short UP~$\omega$-words?
    \item\textbf{RQ2}: Do structural properties of the underlying DBAs influence (a) generalization performance and (b) learned model complexity?
\end{description}

To answer these questions, we use the $\omega$-regular languages associated with two well-known LTL benchmarks. The first benchmark (\textbf{alaska\_lift})~\cite{alaska_tool_citation} consists of two encodings of safe lift behaviors (we use the variants with bug-fixes~\cite{schuppan_lift_fix}) parameterized by the number of floors: encoding (a) uses a linear number of propositions per floor and encoding (b) uses a logarithmic number of propositions per floor.
The second benchmark (\textbf{acacia\_example})~\cite{acacia_tool_citation} consists of 25 formulas specifying the behavior of arbiters and traffic light controllers~\cite{schuppan_benchmark_table}. For this benchmark, we use the negated versions of the formulas, which in our setting only flip the labels of the generated sequences.

For each LTL formula, we generate its corresponding DBA using Spot~\cite{spot_tool_citation}. During training, we generate sequences on-the-fly with uniformly sampled lengths between $2$ and $64$. Once sampled, we transform sequences into one-hot encoded symbol embeddings for RNN processing. Our test data consists of $512$ sampled sequences for each length between $2$ and $512$. We apply a 10-minute timeout for the construction of each DBA. We excluded automata with more than 200 states due to computationally expensive sequence generation, not RNN training. This constraint caused the exclusion of lift formulas for $4$ or more floors and $2$ formulas from \textbf{acacia\_example}. The test data was balanced for the lift formulas and for $20$ of the \textbf{acacia\_example} formulas. However, the training data was not balanced for the lift formulas and the $3$ \textbf{acacia\_example} formulas with unbalanced test data.

Each experiment trains a single-layer vanilla RNN~\cite{elman1990finding} with a hidden dimension of $256$ (consistent with~\cite{deletang2023neural}), batch size $256$ for $1\times10^5$ steps. We use linear warmup for the learning rate~\cite{linear_warmup} from $1\times10^{-8}$ to $1\times 10^{-3}$ at $20\%$ of the training steps, the AMSGrad optimizer~\cite{amsgrad}, and $L^2$ regularization with weight $5\times 10^{-4}$. We minimize cross-entropy loss. Results are from a single seed due to computational budget. All experiments were conducted using an NVIDIA RTX 6000 Ada GPU. We measure in-distribution (ID) accuracy, the mean accuracy of the test data in the training length range, and out-of-distribution (OOD) accuracy, the mean accuracy of the test data outside of the training length range.

To address RQ1, we assess whether RNNs can learn to classify $\omega$-words by measuring performance on the test data, reporting ID accuracy (lengths 2-64) and OOD accuracy (lengths 65-512) in a summary table. To address RQ2a, we plot the relationship between the number of states in the DBAs and generalization performance. To address RQ2b, we plot the number of states against the parameter norm ($L^2$ norm) of the trained models (post-training). We report the Pearson correlation coefficients~\cite{freedman2007statistics} and their statistical significance between the number of states and both OOD accuracy and the trained models' parameter norms to assess their linear correlations.

\subsection{Results}

In Table~\ref{tab:generalization_summary}, we present the results answering RQ1. The majority of tasks ($92.6\%$) had perfect or near-perfect generalization. Two tasks had OOD accuracies of $81.5\%$ and $76.8\%$.

\begin{table}[!htb]
\centering
\caption{Summary of RNN generalization performance on $\omega$-regular language recognition tasks.}
\label{tab:generalization_summary}
\begin{tabular}{@{}lrrrr@{}}
\toprule
\textbf{Performance Category} & \textbf{Tasks} & \textbf{Proportion} & \textbf{Mean ID Acc.} & \textbf{Mean OOD Acc.} \\
\midrule
Perfect ($>99.9\%$) & $23$ & 85.2\% & 100.0\% & 100.0\% \\
Near-Perfect ($98-99.9\%$) & $2$ & 7.4\% & 100.0\% & 99.8\% \\
Good ($95-98\%$) & $0$ & 0\% & -- & -- \\
Moderate ($90-95\%$) & $0$ & 0\% & -- & -- \\
Poor ($<90.0\%$) & $2$ & 7.4\% & 100.0\% & 79.1\% \\
\midrule
\textbf{Overall} & \textbf{27} & \textbf{100\%} & \textbf{100.0\%} & \textbf{98.4\%} \\
\bottomrule
\end{tabular}
\end{table}

\begin{figure}[!htb]
    \begin{subfigure}[t]{0.47\textwidth}
        \centering
        \includegraphics{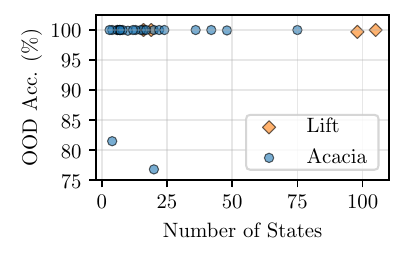}
        \caption{OOD accuracy (acc.) remains consistently high across DBAs with large numbers of states.}
        \label{fig:rq2_left}
    \end{subfigure}\quad
    \begin{subfigure}[t]{0.47\textwidth}
        \centering
        \includegraphics{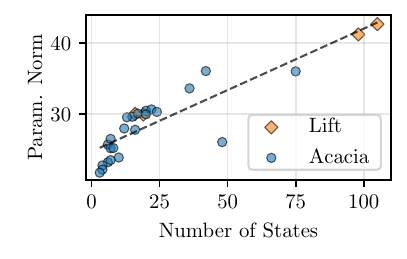}
        \caption{Parameter (param.) norm of the trained RNNs shows a strong positive correlation with the number of states in the DBAs.}
        \label{fig:rq2_right}
    \end{subfigure} %
    \caption{Diamond and circle markers distinguish between \textbf{alaska\_lift} and \textbf{acacia\_example} benchmarks.}
    \label{fig:rq2_analysis}
\end{figure}

In Figure~\ref{fig:rq2_analysis}, we present the plots comparing the number of states with OOD accuracy (Figure~\ref{fig:rq2_left}) and the models' parameter norms (Figure~\ref{fig:rq2_right}). Figure~\ref{fig:rq2_left} reveals that generalization performance remains consistently high across automata of varying complexity, with $85.2\%$ of tasks achieving perfect accuracy regardless of state count, indicating that RNN capacity is robust to the structural complexity of the underlying BA. OOD accuracy showed no significant correlation with the number of states in the DBA ($r = 0.115$, $p=0.567$), suggesting that generalization performance does not necessarily degrade with increasing DBA complexity. Figure~\ref{fig:rq2_right} shows a strong positive correlation between the number of states in the DBA and the parameter norm of the trained models ($r = 0.880, p < 0.001$), indicating that model complexity aligns with the complexity of the target $\omega$-regular languages.

\begin{figure}[!htb]
    \begin{subfigure}[t]{0.47\textwidth}
        \centering
        \includegraphics{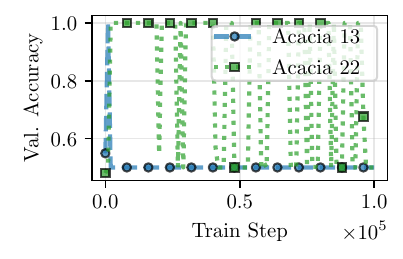}
        \caption{Each model's validation (val.) accuracy on a set of $1024$ sampled sequences of length $512$ during training.}
        \label{fig:rq3_left}
    \end{subfigure}\quad
    \begin{subfigure}[t]{0.47\textwidth}
        \centering
        \includegraphics{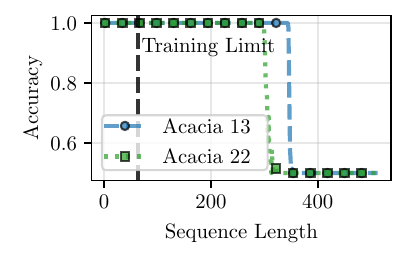}
        \caption{Range evaluation displaying each model's accuracy on test data across sequence lengths.}
        \label{fig:rq3_right}
    \end{subfigure}
    \caption{Validation accuracy and length generalization for tasks with poor generalization.}
    \label{fig:rq3_metrics}
\end{figure}

We now examine the two tasks that exhibited poor generalization to understand their failure modes. During training, both tasks demonstrate unstable validation patterns that predict their generalization failures (see Figure~\ref{fig:rq3_left}). \emph{Acacia 13} achieves $100\%$ validation accuracy initially but then degrades to random chance ($50\%$), while \emph{Acacia 22} oscillates between $100\%$ and $50\%$ throughout training, indicating that neither model converges to a stable representation of the ground truth $\omega$-regular language. This instability corresponds to poor length generalization, as shown in Figure~\ref{fig:rq3_right}, where both tasks achieve perfect accuracy on shorter OOD sequences before experiencing sharp degradation to $50\%$ at longer lengths (degradation begins at $345$ and $300$ for \emph{Acacia 13} and \emph{Acacia 22}, respectively). This degradation suggests that these models converged to local minima, achieving perfect in-distribution accuracy without learning the ground truth $\omega$-regular language. The DBAs of both failure cases contained accepting sink states. If a sequence reaches such a state, the RNN must encode this information when processing all subsequent symbols. Further investigation is needed to confirm this hypothesis.

\section{Conclusion and Future Work}

Our experiments demonstrate that RNNs generalize to the recognition of $\omega$-regular languages from short UP~$\omega$-words. Across 27 tasks with a diverse range of system behaviors, we achieved perfect or near-perfect generalization in $92.6\%$ of cases when testing on sequences up to $8$ times longer than training examples. This finding remained consistent across automata ranging from $3$ to $105$ states, demonstrating that the approach scales to realistic verification problems. These results provide a foundation for developing \emph{differentiable Büchi automata}, components within neurosymbolic systems. These components might behave as monitors in reinforcement learning or enable gradient-based search in model checking.

This work has several limitations that future research should address. Sampling sequences from complex automata (>200 states) proved impractical despite our attempts to speed up the process. Our experiments were also restricted to DBAs as a first step; notably, DBAs cannot represent all $\omega$-regular languages~\cite{Baier2008}. Further investigation of the failure cases is necessary to develop improved training methods. Despite these limitations, this work provides strong evidence that neural approaches could further enhance neurosymbolic methods by offering more scalable, differentiable alternatives to traditional automata-theoretic methods. Exploring interpretability by adapting existing automata extraction techniques (e.g. \cite{merrill_extracting_2022,weiss_extracting_2024}) may enable verification of the learned representations required for safety-critical applications. 

\begin{acknowledgments}
    This work was supported by the UK EPSRC grant 2760033. The authors would like to thank Frederik Kelbel for reading the paper and the reviewers for their constructive feedback.
\end{acknowledgments}

\section*{Declaration on Generative AI}
During the preparation of this work, the authors used Claude Sonnet 4 in order to: Paraphrase and reword. After using this tool, the authors reviewed and edited the content as needed and take full responsibility for the publication's content. 
\bibliography{references}


\end{document}